%% file: arxiv.tex
\documentclass[runningheads]{llncs}
\usepackage[table,xcdraw]{xcolor}
\usepackage{graphicx}

\usepackage{wrapfig}
\usepackage{subfigure}
\usepackage{makecell}
\usepackage{arydshln}
\usepackage{booktabs}
\usepackage{multirow}
\usepackage{tikz}
\usepackage{comment}
\usepackage{amsmath,amssymb} 
\usepackage{color}
\usepackage[pagebackref=true,breaklinks=true,colorlinks,citecolor=brown]{hyperref}

\usepackage[accsupp]{axessibility}  

\newcommand{\statement}[1]{\noindent\textbf{#1}}

\begin{document}
\pagestyle{headings}
\mainmatter
\def\ECCVSubNumber{940}  

\title{FashionViL: Fashion-Focused Vision-and-Language Representation Learning}

\titlerunning{FashionViL: Fashion-Focused V+L Representation Learning}
%
\author{
Xiao Han\inst{1,2}
\and
Licheng Yu\inst{3}
\and
Xiatian Zhu\inst{1,4}
\and
Li Zhang\inst{5}
\\
Yi-Zhe Song\inst{1,2}
\and
Tao Xiang\inst{1,2}
}
\authorrunning{X. Han et al.}
%
\institute{
Centre for Vision, Speech and Signal Processing (CVSSP), University of Surrey
\and
iFlyTek-Surrey Joint Research Centre on Artificial Intelligence
\and
Meta AI
\and
Surrey Institute for People-Centred Artificial Intelligence, University of Surrey
\and
School of Data Science, Fudan University\\
\email{\{xiao.han,xiatian.zhu,y.song,t.xiang\}@surrey.ac.uk}\\
\email{lichengyu@fb.com} \quad
\email{lizhangfd@fudan.edu.cn} \\
}
\maketitle

\input{Sections/0-abstract}
\input{Sections/1-introduction}
\input{Sections/2-related_work}
\input{Sections/3-methodology}

\input{Sections/4-experiments}
\input{Sections/5-conclusion}

\section*{\centering{\vspace{15mm} \\ \Large \textbf{FashionViL: Fashion-Focused Vision-and-Language Representation Learning \\ \vspace{4mm} -- \\ \vspace{4mm} \large Supplementary Material \\ [40pt]}}}

\input{Sections/a-supp}

\clearpage
\bibliographystyle{splncs04}
\bibliography{egbib}

\end{document}

%% file: Sections/0-abstract.tex
\begin{abstract}
Large-scale Vision-and-Language (V+L) pre-training for representation learning has proven to be effective in boosting various downstream V+L tasks. 
However, when it comes to the fashion domain, existing V+L methods are inadequate as they overlook the unique characteristics of both the fashion V+L data and downstream tasks.  
In this work, we propose a novel {\em fashion-focused} V+L representation learning framework, dubbed as FashionViL.
It contains two novel fashion-specific pre-training tasks designed particularly to exploit two intrinsic attributes with fashion V+L data.
First, in contrast to other domains where a V+L data point contains only a single image-text pair, there could be multiple images in the fashion domain. 
We thus propose a Multi-View Contrastive Learning task for pulling closer the visual representation of one image to the compositional multimodal representation of another image+text.
Second, fashion text (\textit{e.g.}, product description) often contains rich fine-grained concepts (attributes/noun phrases). 
To exploit this, a Pseudo-Attributes Classification task is introduced to encourage the learned unimodal (visual/textual) representations of the same concept to be adjacent.
Further, fashion V+L tasks uniquely include ones that do not conform to the common one-stream or two-stream architectures (\textit{e.g.}, text-guided image retrieval). 
We thus propose a flexible, versatile V+L model architecture consisting of a modality-agnostic Transformer so that it can be flexibly adapted to any downstream tasks. 
Extensive experiments show that our FashionViL achieves new state of the art across five downstream tasks.
{Code is available at \url{https://github.com/BrandonHanx/mmf}.}
\keywords{Vision and Language, Representation learning, Fashion.}
\end{abstract}

%% file: Sections/1-introduction.tex
\section{Introduction}\label{sec:intro}
\input{Figures/0-fashion_data}
Recently, Vision-and-Language (V+L) pre-training has received increasing attention \cite{li2019visualbert,tan2019lxmert,lu2019vilbert,su2019vlbert,chen2020uniter,li2020oscar,radford2021clip,kim2021vilt,li2021albef,wang2021simvlm}. The objective is to learn multimodal representations from large-scale image-text pairs, in order to improve various downstream unimodal or multimodal tasks. These models have proven to be highly effective thanks to two main factors: ($i$) there are plenty of image-text pairs on the Web providing abundant training data for free (no additional annotation required), and ($ii$) Transformer-based model architectures have been widely used to learn the contextualized representation of multimodal inputs. 

In this work, we focus on the fashion domain, for which V+L pre-training seems particularly suitable. First, fashion V+L data are not just copious in volume but also high in quality. Online fashion shopping is increasingly ubiquitous; on an e-commerce website, each product detail page (PDP) contains product images and text, both are of very high quality (\textit{i.e.}, often generated by domain experts). Second, there are plenty of downstream tasks, more so than other domains, in real-world applications, ranging from multimodal product understanding \cite{liao2018interpretable,ma2017towards}, cross-modal retrieval \cite{gao2020fashionbert}, to text-guided image retrieval \cite{wu2021fashioniq}. However, when applied to the fashion domain, we observe that existing SOTA V+L pre-training methods \cite{gao2020fashionbert,zhuge2021kaleido} are less effective compared to other domains (see Sec.~\ref{sec:experiments}). We believe that this is because they are not designed to exploit some unique characteristics of both fashion V+L data and downstream tasks.  

In particular, in  most existing generic domain V+L datasets (\textit{e.g.}, COCO~\cite{lin2014coco} and Flickr30k \cite{plummer2015flickr30k}), each data point is a single image-text pair and the text is often brief (\textit{e.g.}, an image caption as shown in Fig.~\ref{fig:fashion_data}). In contrast, fashion datasets are collected mostly from PDPs on e-commerce sites and thus have two specialties: ($i$) There are typically more than one image associated with a given text. One example is shown in Fig.~\ref{fig:fashion_data}.
The garment `maxi dress' is presented with three different views so that online shoppers can view the dress from different angles. ($ii$) There are many more fine-grained concepts in the text description as the text serves as the product description. As shown in Fig.~\ref{fig:fashion_data}, the fashion text is more focused on the garment itself with very detailed adjectives and nouns, describing its appearance in the title, style, and description.
To show that this is statistically true, 
we calculate the ratio on four combined fashion datasets \cite{rostamzadeh2018fashiongen,han2017fashion200k,yang2020facad,vasileva2018polyvoreoutfits} and two combined generic datasets \cite{plummer2015flickr30k,lin2014coco}.
We found that 82\% of the words in the fashion captions are adjectives or nouns, while this ratio becomes only 59\% for the generic captions. None of the existing V+L models are capable of exploiting these specialties in fashion data. 

Fashion downstream tasks are also more diverse than those in the generic domain, posing a challenge to the V+L pre-training model architecture design. More specifically, in the generic V+L domain, existing models are either single-stream or two-stream, depending on the intended downstream tasks.
For example, the single-stream model~\cite{li2019visualbert,su2019vlbert,chen2020uniter,kim2021vilt,huang2020pixelbert} that operates on the concatenation of image and text tokens are suitable for multimodal fusion tasks such as VQA~\cite{antol2015vqa}, VCR~\cite{zellers2019vcr} and RefCOCO~\cite{yu2016modeling}.
In contrast, the two-stream model~\cite{lu2019vilbert,tan2019lxmert,jia2021align,radford2021clip,sun2021lightningdot} are typically designed for efficient cross-modal retrieval tasks\footnote{A single-stream model can also be applied but it needs to traverse every pair of query and gallery item, resulting in unacceptable retrieval speed in large-scale applications.}. 
However, in the fashion domain, apart from image-text fusion and cross-modal retrieval downstream tasks, there are also tasks for which neither single-stream nor two-stream architectures are suitable.
For example, the text-guided image retrieval task \cite{vo2019tirg,wu2021fashioniq,han2022uigr} not only requires a high-quality fusion of the reference image and the modified text but also an efficient matching between the fused multimodal representation and the candidate image.
Due to the diversity of fashion downstream tasks, the existing models, either one-stream or two-stream, do not have the required flexibility and versatility.

To overcome the limitations of existing models for fashion,  we introduce a novel fashion-focused V+L
representation learning framework termed FashionViL.
Two fashion-focused pre-training tasks are proposed to fully exploit the specialties of fashion data.
The first task is Multi-View Contrastive Learning (MVC).
Given a fashion data item with multiple images/views and one text description, we assume that each modality (no matter it is unimodal or multimodal) should be semantically similar to each other since they are all referring to the same product.
Thus, other than the common image-text matching, we propose to minimize the distance between ($a$) the multimodal representation of one of its views and text, and ($b$) the other views.
The second task is Pseudo-Attributes Classification (PAC), designed to exploit the rich fine-grained fashion concepts in the description.
Specifically, we extract those common attributes/noun phrases from the fashion datasets and construct a pseudo attribute set.
The model then learns to predict those attributes during pre-training explicitly.
PAC encourages the fashion items with the same attribute(s) to be clustered together so that the learned representations become more discriminative. We show that (see Sec~\ref{sec:ablation_study}) these new pre-training tasks are effective and complementary to conventional pre-training tasks such as Image-Text Contrastive Learning (ITC) and Masked Language modeling (MLM).

Furthermore, a flexible and versatile model architecture is designed to make the pre-trained model easily adaptable to a diverse set of downstream tasks. The new design keeps the superior fusion ability of single-stream model and the scalability of two-stream model. Crucially, it also caters for fashion-domain unique tasks such as text-guided image retrieval and outfit complementary item retrieval.  
Specifically, our model consists of an image encoder and a modality-agnostic Transformer module, which can be used as either a text encoder or a multimodal fusion encoder.
It thus can be easily fine-tuned for three different downstream use cases: ($i$) early-fusion single-stream mode for joint representation learning, \textit{e.g.}, multimodal classification; ($ii$) late-fusion two-stream mode for unimodal representation learning, \textit{e.g.}, cross-modal retrieval; ($iii$) early-fusion two-stream architecture for compositional representation learning, \textit{e.g.}, text-guided image retrieval. 

In summary, our contributions are as follows: 
(1) A novel V+L pre-training framework is proposed specifically for the fashion domain, which can exploit the specialties of fashion data through two new V+L pre-training tasks.
(2) A flexible architecture design is introduced with a shared text encoder and fusion encoder, which can be easily adapted to a set of diverse fashion downstream tasks.
(3) To demonstrate the generalization of FashionViL, we evaluate our model on 5 fashion V+L tasks: image-to-text retrieval, text-to-image retrieval \cite{rostamzadeh2018fashiongen}, text-guided image retrieval \cite{wu2021fashioniq}, (sub)category recognition \cite{rostamzadeh2018fashiongen} and outfit complementary item retrieval \cite{vasileva2018polyvoreoutfits}.
The experiments show that FashionViL achieves a new state of the art (SOTA) with a consistent and significant performance boost across every downstream task.
To the best of our knowledge, this is the first work capable of addressing 5 diverse fashion tasks together.

%% file: Figures/0-fashion_data.tex
\begin{figure}[t]
\begin{center}
\includegraphics[width=\linewidth]{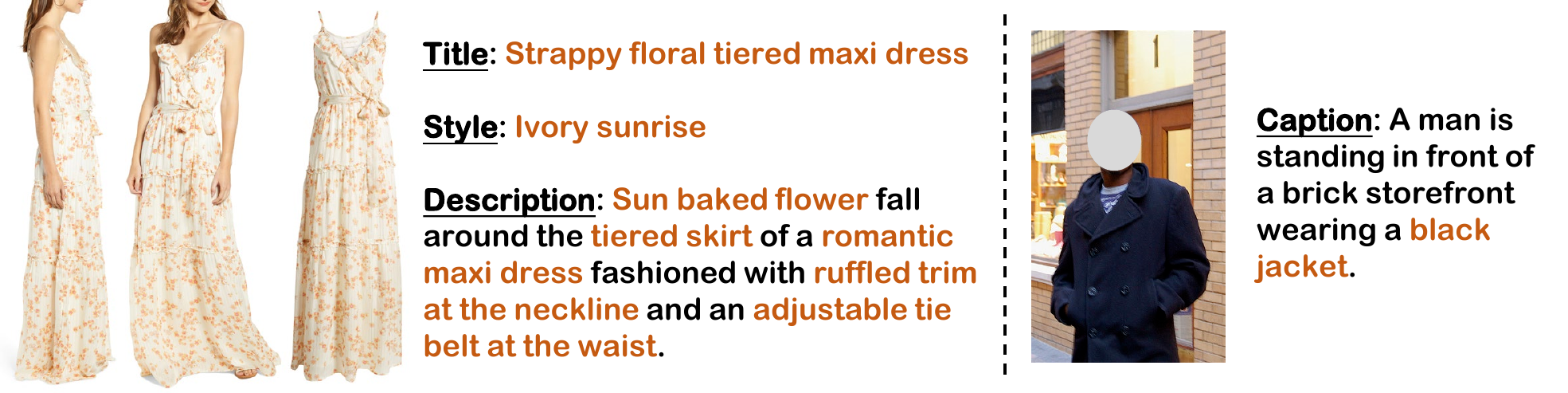}
\end{center}
\caption{Left and right are examples from fashion dataset FACAD \cite{yang2020facad} and Flickr30k \cite{plummer2015flickr30k}, respectively. 
It can be seen that fashion data often have multiple images from different angles, associated with structured titles and descriptions with multiple fine-grained attributes (highlighted in color)}
\label{fig:fashion_data}
\end{figure}

%% file: Sections/2-related_work.tex
\section{Related work}
With the advent of Transformer \cite{vaswani2017transformer} and its success in NLP \cite{devlin2018bert} and CV \cite{dosovitskiy2020vit}, there has been great success in applying large-scale V+L pre-training to generic domain \cite{li2019visualbert,chen2020uniter,li2021albef,radford2021clip}. Some recent studies started to focus on e-commerce domains including fashion
\cite{gao2020fashionbert,zhuge2021kaleido,zhu2021k3m,dong2021m5product,zhang2021ufcbert}. Existing works differ in two main aspects: architecture design and pre-training tasks.

\noindent \textbf{Model architecture.} All V+L pre-training methods use image and text embedding sequences as input for modeling inter-modal and optionally intra-modal interactions through a CNN or Transformer architecture, and output a contextualized feature sequence \cite{bugliarello2021volta}.
There are many options on architecture designs on different aspects, including singe-stream early fusion \cite{li2019visualbert,su2019vlbert,chen2020uniter,li2020oscar} \textit{vs.} two stream late fusion \cite{tan2019lxmert,lu2019vilbert,jia2021align,radford2021clip,fei2021wenlan}, or different visual features (\textit{e.g.,} detector-based regions \cite{zhang2021vinvl} \textit{vs.} ConvNet  patches \cite{huang2020pixelbert} \textit{vs.} linear projections \cite{kim2021vilt,xu2021e2evlp}). In many case, the design is driven by the intended downstream tasks (\textit{e.g.}, VQA requires earlier fusion to enhance joint representation whereas cross-modal retrieval requires later fusion to speed up inference). There are also efforts for alleviating the gap between different architectures through retrieve-and-rerank strategy \cite{sun2021lightningdot,geigle2021mmt-retrieve} or knowledge distillation \cite{wang2021distilled,liu2021inflate}.
Unlike them, inspired by the recent advances in modality-agnostic models \cite{akbari2021vatt,you2021ma-clip,wang2021vlmo,wang2021ufo,li2021unsupervisedvisualbert}, we introduce a unified architecture that can be easily switched between the single-stream or two-stream mode, so there is no need to modify the architecture for different downstream tasks.

\noindent \textbf{Pre-training tasks.} Various tasks have been proposed for V+L pre-training.
Masked Language Modeling (MLM) and Image-Text Matching (ITM) are the direct counterparts of the BERT objectives \cite{devlin2018bert,li2019visualbert}.
Masked Image Modeling (MIM) is the extension of MLM on the visual modality, including several variants like masked region classification \cite{lu2019vilbert,su2019vlbert} and masked region feature regression \cite{chen2020uniter}.
Some other tasks are also proved to be effective, such as predicting object tags \cite{li2020oscar,hu2021vivo}, sequential caption generation \cite{zhou2020unifiedvlp,wang2021simvlm} and image-text contrastive learning \cite{li2021albef,radford2021clip,li2021unimo}.
However, none of these tasks are able to take advantage of the two specialities of fashion data as discussed earlier.
We therefore propose two fashion-focused pre-training tasks in this work.

%% file: Sections/3-methodology.tex
\section{Methodology}
\subsection{Model overview}
\label{sec:model_overview}
\input{Figures/1-2-architecture_and_pretrain}
The model architecture of FashionViL is illustrated in Fig.~\ref{fig:model_architecture_with_pretrain}(a), which is composed of an image encoder (IE) and a Transformer module that can be used for both text encoder (TE) and fusion encoder (FE).
Specifically, our image encoder uses ConvNet as its backbone to convert the raw pixels into a sequence of visual embeddings by rasterizing the grid features of the final feature map. 
For the text encoder, we follow BERT~\cite{devlin2018bert} to tokenize the input sentence into WordPieces~\cite{wu2016wordpieces}.
Each sub-word token's embedding is obtained by summing up its word embedding and learnable position embedding, followed by LN~\cite{ba2016ln}.

One novelty of the model design lies in the shared Transformer for TE and FE, which allows us to flexibly build various multimodal model architectures, each of which is suited for different types of downstream tasks.
For example, Fig.~\ref{fig:model_architecture_with_pretrain}(b) shows an early-fusion model architecture, where the raw sentence and the computed image embeddings are jointly fed into the multimodal fusion encoder. 
Note that when we use the Transformer as the fusion encoder, we will further add the modality embeddings to the visual embeddings and word embeddings, helping the model distinguish the modality type.
This architecture is exactly the same as the well-known single-stream models in many previous pre-training works~\cite{li2019visualbert,chen2020uniter,gao2020fashionbert}.
Then in Fig.~\ref{fig:model_architecture_with_pretrain}(c) we show a late-fusion two-stream model architecture, where we apply the shareable Transformer as the text encoder.
The outputs from image encoder and text encoder are interacted with a simple dot product to compute the similarity between two modalities.
This architecture has been widely adopted for efficient large-scale cross-modal retrieval \cite{sun2021lightningdot,geigle2021mmt-retrieve}.
Furthermore, we can fine-tune this shared Transformer to a more complicated two-stream architecture variant, shown in Fig.~\ref{fig:model_architecture_with_pretrain}(d). 
Here, one stream operates in an early-fusion manner while the other stream is an image encoder.
This architecture is needed for some fashion-focused retrieval tasks with multimodal query, \textit{e.g.}, text-guided image retrieval \cite{vo2019tirg,wu2021fashioniq}.
Note that all FE and TE in the above three architectures are actually the same Transformer, and the mere difference lies in its input.

Given an image-text pair, we denote its raw visual inputs as $\mathbf{v}_{i} = \left\{\mathbf{v}_{i}^{1}, \ldots, \mathbf{v}_{i}^{K}\right\}$, and its input words as $\mathbf{w}_{i}=\left\{\mathbf{w}_{i}^{\mathrm{cls}}, \mathbf{w}_{i}^{1}, \ldots, \mathbf{w}_{i}^{T}\right\}$,
where the subscript $i$ indicates the i-th pair in the dataset. 
An additional special \texttt{[CLS]} token is inserted at the beginning of the text sequence, as well as the multimodal sequence when modalities are concatenated.
We follow the common pre-training + fine-tuning pipeline when applying the model to downstream tasks.

\subsection{Pre-training tasks}
We first introduce two new pre-training tasks. 
This is followed by the other conventional pre-training tasks adopted in our framework. 

\noindent \textbf{Multi-view contrastive learning (MVC).}
As can be seen in Fig.~\ref{fig:fashion_data}, each fashion item is often associated with multiple views to provide a comprehensive overview of the product.
To take advantage of the reciprocal information between different views, we propose to build a correlation between (a) the visual representation of the original view $\mathbf{v}$, and (b) the compositional representation of another view $\mathbf{d}$ and the text $\mathbf{w}$.
In cases where there is only one view of the product, we augment another view by randomly cropping or horizontally flipping the given view.
As shown in Fig.~\ref{fig:model_architecture_with_pretrain}(d), the visual representation of the original view is extracted by the image encoder while the compositional representation is calculated in an early fusion way.
Therefore, the similarity between the multimodal input $[\mathbf{w};\mathbf{d}]$\footnote{We randomly dropout some words in $\mathbf{w}$ and patches in $\mathbf{d}$ with the probability of 15\% to make the learning process more robust.} and $\mathbf{v}$ can be computed as:
\begin{equation}
s\left([\mathbf{w}_{i};\mathbf{d}_{i}], \mathbf{v}_{j}\right)=g_{\theta}\left(\mathbf{d}_{i}^{\mathrm{avg}}|\mathbf{w}_{i}\right)^{T} g_{\theta}\left(\mathbf{v}_{j}^{\mathrm{avg}}\right),
\end{equation}
where $g$ represents a linear transformation that projects the average pooled features into the normalized low-dimensional latent space.
Next, we apply two symmetrical InfoNCE losses \cite{oord2018infonce} to pull closer the matched compositional representations and visual representations in the shared latent space:
\begin{equation}
\mathcal{L}_{\mathrm{InfoNCE}}(x, y)=-\mathbb{E}_{(x, y) \sim B} \log \frac{\exp (s(x, y) / \tau)}{\sum_{\hat{y} \in \hat{B}} \exp (s(x, \hat{y}) / \tau)},
\label{eqa:infonce}
\end{equation}
\begin{equation}
    \mathcal{L}_{\mathrm{MVC}} = \frac{1}{2} \left[\mathcal{L}_{\mathrm{InfoNCE}}([\mathbf{w};\mathbf{d}], \mathbf{v}) + \mathcal{L}_{\mathrm{InfoNCE}}(\mathbf{v}, [\mathbf{w};\mathbf{d}])\right],
\end{equation}
where $\tau$ is a learnable temperature and $\hat{B}$ contains the positive sample $y$ and $|\hat{B}|-1$ negative samples drawn from a mini-batch $B$.

\noindent \textbf{Pseudo-attribute classification (PAC).}
As mentioned in Sec.~\ref{sec:intro}, we found that there are a large number of fine-grained attributes in the fashion description.
We propose to mine the pseudo-attribute concepts from all the available textual information, including title, description and meta-info.
Specifically, we extract all nouns and adjectives via NLTK tagger \cite{bird2009nltk} and only keep those that appear more than 100 times, resulting in a list of 2,232 attributes.
We show the histogram of the top-50 pseudo attributes in Fig. \ref{fig:attribute_histogram}.
It is observed that all of them are truly highly-related to the fashion domain.

Then we explore how to utilize such mined concepts.
We aim to let our model learn to explicitly recognize those pseudo attributes during the pre-training stage.
We model this task as a multi-label classification problem, called Pseudo-Attribute Classification (PAC).
As shown in Fig.~\ref{fig:model_architecture_with_pretrain}(c), we apply the PAC to both visual and textual modalities so that both encoders can learn to capture the fine-grained concepts. 
As this is a weakly-supervised learning setting, we leverage label smoothing to generate the labels \cite{hoe2021orthohash} considering that the mined labels can be noisy.
We use $A$ to denote the whole 2,232 pseudo-attribute set and $a$ as the smoothed soft-target for each class.
For example, if one sample has two ground truth labels at position $0$ and $1$, then $a_0 = a_1 = 0.5$ while $a_i = 0 \ (i \neq 0, 1)$.
Our objective is as follows:
\begin{equation}
    \mathcal{L}_{\mathrm{PAC}}=-\mathbb{E}_{(\mathbf{w}, \mathbf{v}) \sim D} \mathbb{E}_{a \sim A} \left[a \log P_{\theta}\left(a|\mathbf{w}\right) + a \log P_{\theta}\left(a|\mathbf{v}\right)\right],
\end{equation}
where $\theta$ is the learnable parameters and each pair $(\mathbf{w}, \mathbf{v})$ is sampled from the whole training set $D$.

\noindent \textbf{Masked patch feature classification (MPFC).}
While the naive masked feature regression has been shown not helpful in V+L pre-training~\cite{kim2021vilt,dou2021meter}, we found empirically our version of masked patch modeling being effective in the fashion domain.
Specifically, we disregard the feature reconstruction of each masked patch, but instead predict the patch label given by an offline image tokenizer.
To this end, we first train a discrete VAE \cite{van2017vqvae,ramesh2021dalle,esser2021taming} as the image tokenizer on our collected fashion images with the perceputal loss \cite{dong2021peco}.
We also adopt exponential moving average (EMA) to update the codebook, which is proved to be useful for increasing the utilization of codewords \cite{van2017vqvae,dong2021peco}.
We randomly replace 25\% patch features with zeros through block-wise masking strategy \cite{bao2021beit}\footnote{Following UNITER, we use conditional masking for MLM/MPFC, \textit{i.e.}, only masking one modality while keeping the other one intact at each time.}.
Since now we have discrete labels for each patch, the model can be trained to predict the label of each masked patches $\mathbf{v_m}$ given the remaining patches $\mathbf{v_{\backslash m}}$ by optimizing:
\begin{equation}
    \mathcal{L}_{\mathrm{MPFC}}=-\mathbb{E}_{(\mathbf{w}, \mathbf{v}) \sim D} \log P_{\theta}\left(\mathbf{v^t_m}|\mathbf{v}_{\backslash \mathbf{m}}, \mathbf{w}\right),
\end{equation}
where $\mathbf{v^t_m}$ is the estimated target label for the masked patch.

\input{Figures/3-attribute_histogram}
\noindent \textbf{Image-text contrastive learning (ITC).}
We also use ITC to encourage the two unimodal representations to be close in the latent space.
As shown in Fig.~\ref{fig:model_architecture_with_pretrain}(c), the similarity of $\mathbf{w}$ and $\mathbf{v}$ is measured by the dot product of their average pooled features after being projected to the latent space with two linear transformations $f$ and $g$:
$
s\left(\mathbf{w}_{i}, \mathbf{v}_{j}\right)=f_{\theta}\left(\mathbf{w}_{i}^{\mathrm{avg}}\right)^{T} g_{\theta}\left(\mathbf{v}_{j}^{\mathrm{avg}}\right).
$
The ITC loss is:
\begin{equation}
    \mathcal{L}_{\mathrm{ITC}} = \frac{1}{2} \left[\mathcal{L}_{\mathrm{InfoNCE}}(\mathbf{w}, \mathbf{v}) + \mathcal{L}_{\mathrm{InfoNCE}}(\mathbf{v}, \mathbf{w})\right].
\end{equation}

\noindent \textbf{Masked language modeling (MLM).}
In MLM, we randomly mask out the input words with a probability of 15\%, and replace all subwords belonging to the masked words $\mathbf{w}_{\mathbf{m}}$ with special token \texttt{[MASK]}\footnote{Following BERT and UNITER, we decompose this 15\% into 10\% random words, 10\% unchanged, and 80\% \texttt{[MASK]}.}.
The goal of MLM is to predict these masked sub-words based on the observation of their surrounding words $\mathbf{w}_{\backslash\mathbf{m}}$ and all image patches $\mathbf{v}$, by minimizing the negative log-likelihood:
\begin{equation}
    \mathcal{L}_{\mathrm{MLM}}=-\mathbb{E}_{(\mathbf{w}, \mathbf{v}) \sim D} \log P_{\theta}\left(\mathbf{w}_{\mathbf{m}}|\mathbf{w}_{\backslash \mathbf{m}}, \mathbf{v}\right).
\end{equation}

\noindent \textbf{Image-text matching (ITM).}
In ITM, the input is an image-text pair and the target is a binary label $z \in \{0, 1\}$, indicating if each input pair is a match. 
Following \cite{li2021albef}, we sample the hard negative pairs from the similarity matrix $s\left(\mathbf{w}_{i}, \mathbf{v}_{j}\right)$ computed by ITC and then make a mini-batch $H$ containing 50\% negative pairs.
We extract the hidden output of \texttt{[CLS]} at the last layer to represent the joint representation of both modalities, then feed it into a FC layer to do a two-class classification.
We apply cross-entropy loss for ITM:
\begin{equation}
    \mathcal{L}_{\mathrm{ITM}}=-\mathbb{E}_{(\mathbf{w}, \mathbf{v}) \sim H} \log P_{\theta}\left(z|\mathbf{w}, \mathbf{v}\right).
\end{equation}

%% file: Figures/1-2-architecture_and_pretrain.tex
\begin{figure}[t]
\begin{center}
\includegraphics[width=\linewidth]{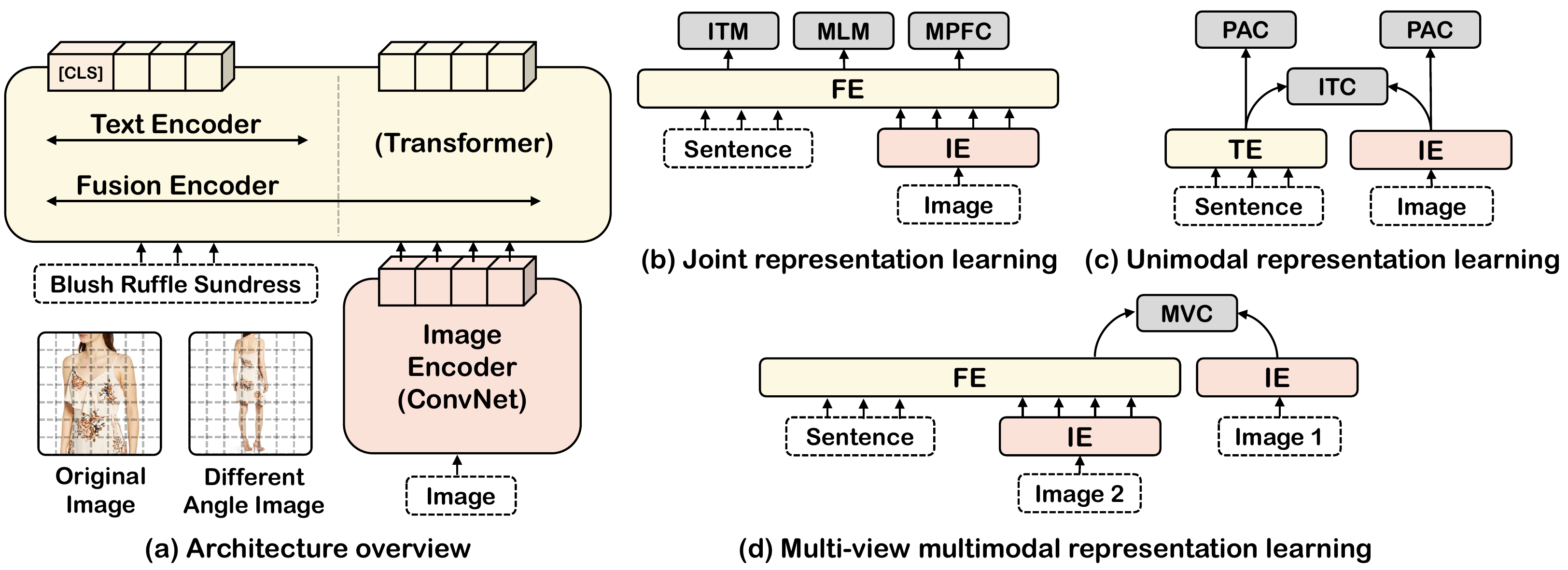}
\end{center}
\caption{Overview of the proposed FashionViL model architecture, consisting of an image encoder, a text encoder and a fusion encoder. Text encoder and fusion encoder share the same parameters. We adopt six pre-training tasks to learn different representations}
\label{fig:model_architecture_with_pretrain}
\end{figure}

%% file: Figures/3-attribute_histogram.tex
\begin{figure}[t]
\begin{center}
\includegraphics[width=\linewidth]{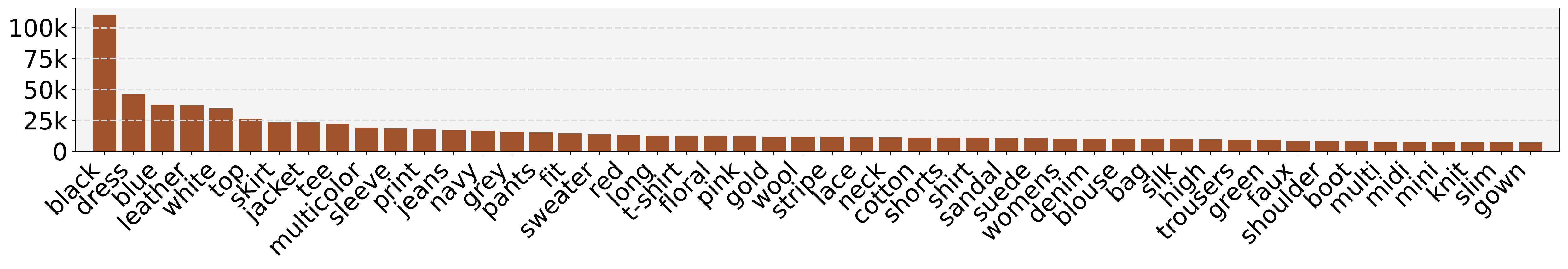}
\end{center}
\caption{Histogram of the top-50 pseudo attributes}
\label{fig:attribute_histogram}
\end{figure}

%% file: Sections/4-experiments.tex
\section{Experiments}
\label{sec:experiments}
In this section, we introduce our pre-training dataset and 5 practical downstream tasks.
We use MMF~\cite{singh2020mmf} and PyTorch \cite{paszke2019pytorch} for the implementation.
For the image encoder, we use an off-the-shelf ResNet50 \cite{he2016resnet} to fairly compare with previous methods, most of which also used ResNet50.
For the text encoder and multimodal fusion encoder (using the shared Transformer), we use the BERT-base-uncased \cite{su2019vlbert} as the initialization.
We use 4 RTX 3090 GPUs for the pre-training.
The details of the hyper-parameters are listed in the supplementary file.

\subsection{Pre-training dataset and downstream tasks}
\input{Tables/0-datasets}
\noindent \textbf{Pre-training dataset.}
Our pre-training dataset consists of 4 public fashion-related datasets, namely, FashionGen \cite{rostamzadeh2018fashiongen}, FACAD \cite{yang2020facad}, Fashion200K \cite{han2017fashion200k} and PolyvoreOutfits \cite{vasileva2018polyvoreoutfits}. 
In total, these datasets provide us with 373.5K fashion products for pre-training.
Because each product may contain multiple images from different angles, we have about 1.35 million image-text pairs on hand.
The detailed statistics are provided in Table \ref{tab:datasets}.

\noindent \textbf{Cross-modal retrieval.}
Image-to-Text Retrieval (ITR) is a cross-modal retrieval task.
Given an image query, our model finds the most aligned text from a large candidate pool.
Previous fashion-domain pre-training works \cite{gao2020fashionbert,zhuge2021kaleido} use the joint representation over the \texttt{[CLS]} token to predict the matching score, which results in an impractical time complexity due to the exhaustive matching between each query item and all gallery items in the early-fusion model~\cite{sun2021lightningdot,wang2021distilled,liu2021inflate,zhang2021vldeformer,geigle2021mmt-retrieve}.
While one of our model architectures can do the same (as Fig.~\ref{fig:model_architecture_with_pretrain}(b)), we opt to use the two-stream late-fusion model in Fig.~\ref{fig:model_architecture_with_pretrain}(c) to compute the cosine similarity for a far more efficient retrieval as~\cite{jia2021align,radford2021clip}.
Text-to-Image Retrieval (TIR) is an inverse problem of ITR, where the query modality and gallery modality are swapped.
The architecture for TIR is the same as ITR.

\noindent \textbf{Text-guided image retrieval (TGIR).}
TGIR is a special type of image retrieval problem, whose query is a multimodal composition~\cite{guo2018dialog,vo2019tirg,wu2021fashioniq,han2022uigr}.
Specifically, given a query image and a modified sentence, the model is required to retrieve another image which has the similar outlook as the query image but with some appearance changes according to the query text.
It has many practical applications in fashion, such as retrieving another garment according to a user's reference garment and his/her feedback.
To handle the uniqueness of the multimodal query, several interesting fusion approaches have been proposed in the past, such as
the gating mechanism \cite{vo2019tirg,shin2021rtic}, hierarchical attention \cite{chen2020val}, and style-content modification \cite{lee2021cosmo}.
In this work, we follow~\cite{liu2021cirr} to simply apply an early fusion model to encode the compositional representation of the query image and modified text, which is shown in Fig.~\ref{fig:model_architecture_with_pretrain}(d).

\noindent \textbf{Category/Subcategory recognition (CR/SCR).}
The (sub)category is a vital attribute for describing a product. 
(S)CR requires the model to produce a reliable joint representation.
Following previous works \cite{gao2020fashionbert,zhuge2021kaleido}, we directly append a linear layer on top of \texttt{[CLS]} to predict the label for these tasks.

\noindent \textbf{Outfit complementary item retrieval (OCIR).}
OCIR aims at finding visually compatible item(s) of several given items to complete an outfit.
This is a very practical task as people often buy garments that match previously selected or purchased ones.
OCIR can be a helpful recommendation feature for online retailers \cite{lin2020csanet,hou2021disentangled}.
To address this task, we replace the backbone of CSA-Net \cite{lin2020csanet} with the pre-trained image encoder of FashionViL.
Note that unlike all multimodal/cross-modal tasks above, only the pre-trained image encoder is used in this downstream task.
We leverage this task to evaluate the performance of our image encoder under the proposed multimodal pre-training.

\subsection{Comparative results}
\noindent \textbf{Cross-modal retrieval.}
\input{Tables/1-xmodal_retrieval}
We evaluate the cross-modal retrieval on the FashionGen \cite{rostamzadeh2018fashiongen} test split (not included in pre-training), including both ITR and TIR.
Table \ref{tab:xmodal_retrieval_results} compares the performance of the previous V+L pre-training methods with our FashaionViL.
Because previous works~\cite{gao2020fashionbert,zhuge2021kaleido} are designed with a single-stream architecture, they can only be evaluated on a small retrieval set.
For example, for TIR, the models are required to pick the best-matched image from only 101 images given a text query\footnote{In the 101 images, 1 is positively paired with the text and the other 100 are randomly paired but sharing the same sub-category as the positive, increasing the difficulty.}. 
Recall (over 1K retrievals) is reported as the metric.
The same setting is used for ITR.
For a fair comparison, we strictly follow the same evaluation protocol, reporting the recall for 1K retrievals\footnote{Because the authors did not release their 1K retrieval set, we report the average recall of 5 experiments with 5 randomly selected 1K retrieval sets.}.

In Table~\ref{tab:xmodal_retrieval_results}, we compare our FashionViL and its two variants with existing methods. In particular,  \textit{-e2e} and \textit{-pt} denotes our model without end-to-end training (image encoder is fixed) and multimodal pre-training respectively. 
We have the following observations: (1) Even with the fixed image encoder and without pre-training, FashionViL already achieves comparable results with the existing  methods. This suggests that the performance of late fusion can be as effective as early-fusion for such fine-grained cross-modal retrieval. 
(2) When we unfreeze the image encoder for end-to-end training, we observe that $R@1$ jumps from $21.13$ to $58.84$, suggesting that end-to-end training is very efficient and redundant pre-processing may be unnecessary.
(3) When we further utilize our proposed multimodal pre-training, our model achieves SOTA performance as in the last column of Table.~\ref{tab:xmodal_retrieval_results}, whose $R@1$ is more than twice of the previous SOTA. 

\input{Tables/1b-xmodal_ours}
Note that our model architecture for this task is two-stream.
This means that it can be applied to large-scale retrieval, unlike the compared baselines.
Therefore, we additionally report the evaluation results on the full test set (of 32K image-text pairs), \textit{i.e.}, each query item is compared with every gallery item in the full test set.
The results can be found in Table~\ref{tab:xmodal_ours}. We encourage the future works to also follow such a full evaluation protocol to measure the performance.

\noindent \textbf{Text-guided image retrieval.}
\input{Tables/2-compositional_retrieval}
\input{Tables/3-it_classification}
For TGIR, we compare our FashionViL with the previous V+L pre-training methods and the task-specific methods on FashionIQ~
\cite{wu2021fashioniq}\footnote{ 
Details for the reproduction of previous methods are in the supplementary file.
}.
The results are shown in Table \ref{tab:compositional_retrieval_results}.
For more comprehensive comparisons, we use two different implementations adopted by previous methods, \textit{i.e.}, training with fixed image encoder \cite{liu2021cirr} or end-to-end training \cite{vo2019tirg,chen2020val,lee2021cosmo}.

We first report the results with the fixed ResNet 152 from Column 1 to Column 4 (C1-C4).
CIRR adopts OSCAR \cite{li2020oscar} as the fusion module and uses the global image features as the input.
We find FashionViL consistently outperforms CIRR with a relative 10\%$\sim$20\% gain with or without the multimodal pre-training (C1 \textit{vs.} C3, C2 \textit{vs.} C4).
This improvement demonstrates that the patch-level features are superior to the global features for the compositional multimodal fusion.
With our proposed pre-training, the performance further improves from 31.78 to 34.19 (C3 \textit{vs.} C4), showing our pre-training also works well on the off-the-shelf fixed image encoder.

We then report the results under the end-to-end training paradigm (C5-C10).
We find that simply replacing GRU with BERT (C5 \textit{vs.} C8) already leads to a 4\% relative gain (from 23.65 to 27.17), indicating the importance of having a higher-quality text encoder.
Additionally, all previous works apply a late interaction between the image embeddings and modified text embeddings with an elaborately designed fusion module, \textit{e.g.}, TIRG~\cite{vo2019tirg}.
We argue that an earlier fusion of the two modalities should result in an even better compositional embedding for the query purpose.
Comparing C9 and C8, our FashionViL without pre-training already outperforms TIRG+BERT, indicating better query multimodal embeddings are learned in our model.
Note that our text encoder and fusion encoder are shared, so FashionViL also saves more training parameters than TIRG+BERT.
With the help of pre-training, our FashionViL achieves the new SOTA result with another significant 11.2\% relative gain (C9 \textit{vs.} C10).

\noindent \textbf{Category / Subcategory recognition.}
Following KaleidoBERT \cite{zhuge2021kaleido}, we evaluate CR and SCR on the FashionGen dataset~\cite{rostamzadeh2018fashiongen}.
The joint representation of the model architecture in Fig.~\ref{fig:model_architecture_with_pretrain}(b) is used to predict the classification score.
The results are shown in Table \ref{tab:it_classification_results}.
Once again, the end-to-end learning and the well-designed fashion-specific pre-training tasks help our FashionViL outperform the two previous works by  significant margins (10.4\% and 3.2\%, respectively).
Furthermore, we also simulate a new task -- multi-image subcategory recognition (M-SCR) to evaluate the performance of FashionViL with multiple input images.
See more results in the supplementary file.

\input{Tables/4-ocir}
\input{Tables/5-ablation_study}
\noindent \textbf{Outfit complementary item retrieval.}
In addition to the aforementioned multimodal and instance-level downstream tasks, we also examine FashionViL on the unimodal outfit-level task, \textit{i.e.}, OCIR.
We compare our model with the previous task-specific methods \cite{lin2020csanet,hou2021disentangled} on the Disjoint split of Polyvore Outfits \cite{vasileva2018polyvoreoutfits}\footnote{
We have no access to the data splits of CSA-Net, so constructed the Polyvore Outfits~\cite{vasileva2018polyvoreoutfits} and reproduced CSA-Net by ourselves according to the original paper \cite{lin2020csanet,hou2021disentangled}.
}.
As shown in Table \ref{tab:ocir_results}, our multimodal pre-training benefits the performance with a 21.0\% improvement, even when only the image encoder is tuned. 

\subsection{Ablation study}
\label{sec:ablation_study}
We analyze the effectiveness of different pre-training tasks and the sharing TE/FE strategy through ablation studies over the aforementioned five downstream tasks.
The complete results are listed in Table \ref{tab:ablation_study}.
In addition to the standard metrics for each benchmark, we use the Meta-sum (sum of all scores across all the benchmarks) as a global metric.

First, we establish a baseline without any multimodal pre-training in Line 0 (L0), \textit{i.e.}, the image/text encoder is initialized with the off-the-shelf ResNet50 or BERT, which is pre-trained in vision-only or language-only domain.

Second, we validate the effectiveness of each pre-training task by their standalone performance, \textit{i.e.}, each time we pick only one task for pre-training.
We show the results of MPFC, MLM, PAC, MVC, ITC in L2, L4, L5, L6 and L7. 
It is clear from Table~\ref{tab:ablation_study} that all of these pre-training tasks can benefit the downstream tasks.
However, we found that a pre-training task tends to be relatively more helpful to downstream tasks of its similar type.
For example, both MPFC (L2) and MLM (L4) are focusing on modeling the cross-modal interaction, thus they bring more gain to SCR but contribute relatively less to ITR and TIR.
In contrast, since ITC (L7) has the same objective with ITR and TIR, it significantly boosts the cross-modal performance.
As for TGIR, it requires not only high-quality compositional representation but also high-quality unimodal representations, thus each of the 5 pre-training tasks have a positive impact.

Third, we validate the effectiveness of the proposed PAC (L5) and MVC (L6).
For PAC, we implement a comparative experiment: MLM only on those pre-defined pseudo-attribute words (L3).
The main difference between L3 and L5 is whether the multi-label supervision is performed on each masked text token or the global representation.
L3 leads to much lower performance than L5, indicating that the supervision of pseudo attributes on the global representation is a better choice.
Interestingly, L3 achieves a comparable result to L4, where each word (including those other than the pseudo attributes) can also be masked.
This means merely masking the fine-grained words is as effective as masking all the words uniformly, which indicates the most important text cues lie in those fine-grained concept words. 
We then verify the superiority of MVC. 
To this end, we add an ablation study that does not utilize multi-angle images (L1), \textit{i.e.}, replacing the sampled different angle image with an augmented version of the original image.
Comparing L1 and L6, we confirm that the improvement of MVC mainly comes from the contrastive learning on the images from different angles.

Next, we study the effect of different combinations of those tasks.
When we add MLM and MPFC to ITC (L8), we observe a gain on Meta-sum, while the performance of ITR and TIR slightly drops.
This is expected as different tasks may provide different update directions for the same parameters, which causes some tasks to overshadow the effects of others.
However, minor conflicts between different tasks can be largely alleviated by employing more tasks.
As shown in L9, the overall performance can be further boosted by adding ITM.
The same happens when we add MVC into them (L10).
When all six tasks are jointly trained (L11), we observe a significant performance gain across all benchmarks.
Notably, the two new fashion-specific tasks of MVC and PAC play the most important roles to achieve the SOTA performance. 

Finally, we demonstrate the superiority of sharing TE and FE. 
We implement a comparative model (L12) with the same pre-training tasks as L11 but using separate TE and FE.
We observe a clear performance drop when breaking the parameter sharing.
This indicates our modality-agnostic sharing strategy not only reduces the number of parameters but also performs far better.

\subsection{Visualization}
\input{Figures/4-tsne_embeddings}
We visualize the representations from the image encoder, text encoder, and fusion encoder via t-SNE \cite{van2008tsne} in Fig.~\ref{fig:tsne_embeddings}.
Specifically, we feed all image-text pairs from FashionGen's test split into our model.
We visualize the most popular 10 categories using different colors.
We compare the t-SNE of the model without multimodal pre-training (initialized with ResNet+BERT) and the model with the full 6 pre-training tasks.
We found the clusters become more discriminative when more pre-training tasks are added, indicating that FashionViL learns to acquire more fine-grained concepts.
See more in the supplementary file.

%% file: Tables/0-datasets.tex
\begin{table}[t]
\caption{Statistics on the datasets used for pre-training}
\label{tab:datasets}
\resizebox{\textwidth}{!}{
\begin{tabular}{c|cccccccc|cc}
\hline
\multirow{2}{*}{\textbf{Datasets}} & \multicolumn{2}{c}{\textbf{FashionGen} \cite{rostamzadeh2018fashiongen}} & \multicolumn{2}{c}{\textbf{FACAD} \cite{yang2020facad}}     & \multicolumn{2}{c}{\textbf{Fashion200k} \cite{han2017fashion200k}} & \multicolumn{2}{c|}{\textbf{PolyvoreOutfits} \cite{vasileva2018polyvoreoutfits}} & \multicolumn{2}{c}{\textbf{Total}}     \\
                                   & \textit{\#products}  & \textit{\#pairs} & \textit{\#products} & \textit{\#pairs} & \textit{\#products}  & \textit{\#pairs}  & \textit{\#products}     & \textit{\#pairs}    & \textit{\#products} & \textit{\#pairs} \\ \hline
\textbf{Train}                     & 60k                  & 260k             & 164.5k              & 847k             & 77k                  & 172k              & 72k                     & 72k                 & 373.5k              & 1.35M            \\
\textbf{Val}                       & 7.5k                 & 32.5k            & 18k                 & 94k              & 13k                  & 30k               & 14.5k                   & 14.5k               & 53k                 & 171k             \\ \hline
\end{tabular}
}
\end{table}

%% file: Tables/1-xmodal_retrieval.tex
\begin{table}[t]
\caption{Results of cross-modal retrieval on FashionGen \cite{rostamzadeh2018fashiongen} with the protocol used in KaleidoBERT \cite{zhuge2021kaleido}. \textit{-e2e}: without end-to-end training, \textit{i.e.}, the image encoder is fixed. \textit{-pt}: directly fine-tuning without multimodal pre-training}
\label{tab:xmodal_retrieval_results}
\resizebox{\textwidth}{!}{
\begin{tabular}{cc|ccccccc|ccc}
\hline
\multicolumn{2}{c|}{\multirow{2}{*}{\textbf{Methods}}}             & \textbf{VSE}++  & \textbf{ViLBERT} & \textbf{VLBERT}  & \textbf{Image-}   & \textbf{Fashion-} & \textbf{OSCAR} & \textbf{Kaleido-} & \multicolumn{3}{c}{\textbf{Ours}}                                                      \\ 
\multicolumn{2}{c|}{}                                              & \cite{vaghri2019vsepp}                          & \cite{lu2019vilbert}     & \cite{su2019vlbert}   & \textbf{BERT} \cite{qi2020imagebert}      & \textbf{BERT} \cite{gao2020fashionbert}   & \cite{li2020oscar}      & \textbf{BERT} \cite{zhuge2021kaleido}    & \multicolumn{1}{c}{\textit{-e2e -pt}} & \multicolumn{1}{c}{\textit{-pt}} & \qquad\qquad\quad   \\ \hline
\multicolumn{1}{c|}{\multirow{3}{*}{\textbf{ITR}}} & \textbf{R@1}  & 4.59                     & 20.97            & 19.26    & 22.76      & 23.96             & 23.39          & 27.99             & \multicolumn{1}{c}{21.13}             & \multicolumn{1}{c}{58.84}  &  \textbf{65.54}    \\
\multicolumn{1}{c|}{}                              & \textbf{R@5}  & 14.99                   & 40.49            & 39.90     & 41.89      & 46.31             & 44.67          & 60.09             & \multicolumn{1}{c}{46.82}             & \multicolumn{1}{c}{89.46}   &  \textbf{91.34}     \\
\multicolumn{1}{c|}{}                              & \textbf{R@10} & 24.10                   & 48.21            & 46.05     & 50.77      & 52.12             & 52.55          & 68.37             & \multicolumn{1}{c}{58.71}             & \multicolumn{1}{c}{95.84}   &  \textbf{96.30}   \\ \hline
\multicolumn{1}{c|}{\multirow{3}{*}{\textbf{TIR}}} & \textbf{R@1}  & 4.60                     & 21.12            & 22.63    & 24.78       & 26.75             & 25.10          & 33.88             & \multicolumn{1}{c}{25.83}             & \multicolumn{1}{c}{57.16}  &  \textbf{61.88}     \\
\multicolumn{1}{c|}{}                              & \textbf{R@5}  & 16.89                   & 37.23            & 36.48     & 45.20      & 46.48             & 49.14          & 60.60             & \multicolumn{1}{c}{51.54}             & \multicolumn{1}{c}{84.34}   &  \textbf{87.32}    \\
\multicolumn{1}{c|}{}                              & \textbf{R@10} & 28.99                   & 50.11            & 48.52     & 55.90      & 55.74             & 56.68          & 68.59             & \multicolumn{1}{c}{63.53}             & \multicolumn{1}{c}{91.90}   &  \textbf{93.22}    \\ \hline
\multicolumn{2}{c|}{\textbf{Mean}}                                 & 15.69                   & 36.36           & 35.47     & 40.22      & 41.89            & 41.92         & 53.25            & \multicolumn{1}{c}{44.59}               & \multicolumn{1}{c}{79.59}      &  \textbf{82.60}   \\ \hline
\end{tabular}
}
\end{table}

%% file: Tables/1b-xmodal_ours.tex

\begin{table}[t]
\centering
\caption{Results of cross-modal retrieval on FashionGen \cite{rostamzadeh2018fashiongen} with full evaluation}
\label{tab:xmodal_ours}
\setlength\tabcolsep{8pt}
\resizebox{0.6\textwidth}{!}{
\begin{tabular}{ccc|ccc|c}
\hline
\multicolumn{3}{c|}{\textbf{ITR}}           & \multicolumn{3}{c|}{\textbf{TIR}}           & \multirow{2}{*}{\textbf{Mean}} \\ \cline{1-6}
\textbf{R@1} & \textbf{R@5} & \textbf{R@10} & \textbf{R@1} & \textbf{R@5} & \textbf{R@10} &                                \\ \hline
42.88        & 71.57        & 80.55         & 51.34        & 75.42        & 84.75         & 67.75                          \\ \hline
\end{tabular}
}
\end{table}

%% file: Tables/2-compositional_retrieval.tex
\begin{table}[t]
\caption{Results of text-guided image retrieval on FashionIQ \cite{wu2021fashioniq}}
\label{tab:compositional_retrieval_results}
\resizebox{\textwidth}{!}{
\begin{tabular}{cc|cccc|cccccc}
\hline
\multicolumn{2}{c|}{\textbf{Image Encoder}}                           & \multicolumn{4}{c|}{\textbf{Fixed ResNet 152}}                                               & \multicolumn{6}{c}{\textbf{ResNet 50}}                                                                                         \\ \hline
\multicolumn{2}{c|}{\textbf{Fusion Module}}                           & \multirow{2}{*}{\textbf{CIRR}\textit{-pt}} & \multicolumn{1}{c|}{\multirow{2}{*}{\textbf{CIRR} \cite{liu2021cirr}}} & \multirow{2}{*}{\textbf{Ours}\textit{-pt}} & \multirow{2}{*}{\textbf{\ \ Ours\ \ }} & \textbf{TIRG} \cite{vo2019tirg} & \textbf{VAL} \cite{chen2020val} & \textbf{CoSMo} \cite{lee2021cosmo} & \multicolumn{1}{c|}{\textbf{TIRG} \cite{vo2019tirg}} & \multirow{2}{*}{\textbf{Ours}\textit{-pt}} & \multirow{2}{*}{\textbf{Ours}}       \\
\multicolumn{2}{c|}{\textbf{Text Encoder}}                            &        & \multicolumn{1}{c|}{}    &          &     & \textbf{GRU \cite{cho2014gru}}  & \textbf{GRU} \cite{cho2014gru} & \textbf{GRU} \cite{cho2014gru}   & \multicolumn{1}{c|}{\textbf{BERT} \cite{su2019vlbert}} &  &  \\
\multicolumn{2}{c|}{}                                    & (1)            & \multicolumn{1}{c|}{(2)}         & (3)               & \multicolumn{1}{c|}{(4)}         & (5)         & (6)        & (7)          & \multicolumn{1}{c|}{(8)}         & (9)               & (10)      \\ \hline
\multicolumn{1}{c|}{\multirow{2}{*}{\textbf{Dress}}}  & \textbf{R@10} & 14.38             & \multicolumn{1}{c|}{17.45}         & 20.97               & \textbf{22.66}         & 23.65         & 26.28        & 24.49          & \multicolumn{1}{c|}{27.17}         & 28.46               & \textbf{33.47}      \\
\multicolumn{1}{c|}{}                                 & \textbf{R@50} & 34.66             & \multicolumn{1}{c|}{40.41}         & 42.64               & \textbf{46.60}         & 49.93         & 50.25        & 51.01          & \multicolumn{1}{c|}{53.25}         & 54.24               & \textbf{59.94}      \\ \hline
\multicolumn{1}{c|}{\multirow{2}{*}{\textbf{Shirt}}}  & \textbf{R@10} & 13.64             & \multicolumn{1}{c|}{17.53}         & 17.62               & \textbf{18.74}         & 21.98         & 21.69        & 18.99          & \multicolumn{1}{c|}{22.28}         & 22.33               & \textbf{25.17}      \\
\multicolumn{1}{c|}{}                                 & \textbf{R@50} & 33.56             & \multicolumn{1}{c|}{38.31}         & 41.32               & \textbf{41.56}         & 46.61         & 45.53        & 43.57          & \multicolumn{1}{c|}{45.58}         & 46.07               & \textbf{50.39}      \\ \hline
\multicolumn{1}{c|}{\multirow{2}{*}{\textbf{Toptee}}} & \textbf{R@10} & 16.44             & \multicolumn{1}{c|}{21.64}         & 21.67               & \textbf{25.29}         & 27.84         & 27.43        & 25.19          & \multicolumn{1}{c|}{27.84}         & 29.02               & \textbf{34.98}      \\
\multicolumn{1}{c|}{}                                 & \textbf{R@50} & 38.34             & \multicolumn{1}{c|}{45.38}         & 46.46               & \textbf{50.28}         & 55.07         & 56.25        & 54.00          & \multicolumn{1}{c|}{57.11}         & 57.93               & \textbf{60.79}      \\ \hline
\multicolumn{2}{c|}{\textbf{Mean}}                                    & 25.17             & \multicolumn{1}{c|}{30.20}         & 31.78               & \textbf{34.19}         & 37.51         & 37.91        & 36.21          & \multicolumn{1}{c|}{38.87}         & 39.67               & \textbf{44.12}      \\ \hline
\end{tabular}
}
\end{table}

%% file: Tables/3-it_classification.tex
\begin{table}[t]
\centering
\caption{Results of category / subcategory recognition on FashionGen \cite{rostamzadeh2018fashiongen}}
\label{tab:it_classification_results}
\resizebox{0.8\textwidth}{!}{
\begin{tabular}{cc|cccc|cc}
\hline
\multicolumn{2}{c|}{\multirow{2}{*}{\textbf{Methods}}}               & \textbf{FashionBERT} & \textbf{ImageBERT} & \textbf{OSCAR} & \textbf{KaleidoBERT} & \multicolumn{2}{c}{\textbf{Ours}} \\
\multicolumn{2}{c|}{}                                                & \cite{gao2020fashionbert} & \cite{qi2020imagebert}  & \cite{li2020oscar}  & \cite{zhuge2021kaleido}                 & \textit{\quad -pt \quad}   & \textit{}        \\ \hline
\multicolumn{1}{c|}{\multirow{2}{*}{\textbf{CR}}}  & \textbf{Acc}    & 91.25                & 90.77              & 91.79          & 95.07                & 97.07          & \textbf{97.48}   \\
\multicolumn{1}{c|}{}                              & \textbf{Macro}$\mathcal{F}$ & 70.50                & 69.90              & 72.70          & 71.40                & 84.72          & \textbf{88.60}   \\ \hline
\multicolumn{1}{c|}{\multirow{2}{*}{\textbf{SCR}}} & \textbf{Acc}    & 85.27                & 80.11              & 84.23          & 88.07                & 91.45          & \textbf{92.23}   \\
\multicolumn{1}{c|}{}                              & \textbf{Macro}$\mathcal{F}$ & 62.00                & 57.50              & 59.10          & 63.60                & 78.13          & \textbf{83.02}   \\ \hline
\multicolumn{2}{c|}{\textbf{Mean}}                                   & 77.76                & 74.57              & 76.96          & 79.54                & 87.84          & \textbf{90.33}   \\ \hline
\end{tabular}
}
\end{table}

%% file: Tables/4-ocir.tex
\begin{table}[t]
\centering
\caption{Results of outfit complementary item retrieval on PolyvoreOutfits  \cite{vasileva2018polyvoreoutfits}}
\label{tab:ocir_results}
\resizebox{0.85\textwidth}{!}{
\begin{tabular}{cc|cccc|ccc}
\hline
\multicolumn{2}{c|}{\multirow{2}{*}{\textbf{Methods}}}              & \textbf{Type-aware} & \textbf{SCE-Net} & \textbf{CSA-Net} & \textbf{ADDE-O} & \textbf{CSA-Net}    & \multicolumn{2}{c}{\textbf{Ours}} \\
\multicolumn{2}{c|}{}                                               &    \cite{vasileva2018polyvoreoutfits}  & \cite{tan2019scenet} & \cite{lin2020csanet}   & \cite{hou2021disentangled}    & \small \textit{reproduced}  & \textit{\quad -pt \quad}   & \textit{}        \\ \hline
\multicolumn{1}{c|}{\multirow{3}{*}{\textbf{OCIR}}} & \textbf{R@10} & 3.66                & 4.41             & 5.93             & 6.18            & 2.69                & 4.38           & \textbf{5.83}    \\
\multicolumn{1}{c|}{}                               & \textbf{R@30} & 8.26                & 9.85             & 12.31            & 13.79           & 6.29                & 10.54          & \textbf{12.61}   \\
\multicolumn{1}{c|}{}                               & \textbf{R@50} & 11.98               & 13.87            & 17.85            & 18.60           & 9.14                & 14.77          & \textbf{17.49}   \\ \hline
\multicolumn{2}{c|}{\textbf{Mean}}                                  & 7.97                & 9.38             & 12.03            & 12.86           & 6.04                & 9.90           & \textbf{11.98}   \\ \hline
\end{tabular}
}
\end{table}

%% file: Tables/5-ablation_study.tex
\begin{table}[t]
\caption{Evaluation on pre-training tasks using ITR, TIR, TGIR, SCR and OCIR as downstream tasks. Each number is the mean value of all metrics for one specific downstream task. Meta-sum stands for the summation of all numbers in each row. The three shades of grey represent the top three results when sharing TE and FE}
\label{tab:ablation_study}
\renewcommand\arraystretch{1.25}
\resizebox{\textwidth}{!}{
\begin{tabular}{cl|ccccc|c}
\hline
\multicolumn{2}{c|}{\textbf{Pre-training Tasks}} & \textbf{\, ITR \,}                           & \textbf{\, TIR \,}                           & \textbf{\, TGIR \,}                          & \textbf{\, SCR \,}                           & \textbf{\, OCIR \,}                          & \textbf{\, Meta-sum \,}                                     \\ \hline
(0)    & \textbf{None}                             & 62.50                                  & 68.09                                  & 39.67                                  & 84.79                                  & 9.90                                   & 265.04                                                \\
(1)    & \textbf{MVC} (use augmented image only)        & 62.85                                  & 68.58                                  & 40.50                                  & 84.86                                  & 9.53                                  & 266.32                                                \\
(2)    & \textbf{MPFC}                             & 62.10                                  & 68.12                                  & 40.22                                  & 86.39                                  & 10.05                                  & 266.88                                                \\
(3)    & \textbf{MLM} (mask attribute words only)   & 62.32                                  & 67.93                                  & 40.46                                  & 85.83                                  & 10.38                                  & 266.92                                                \\
(4)    & \textbf{MLM}                              & 62.15                                  & 67.43                                  & 40.29                                  & 86.72                                  & 10.38                                  & 266.97                                                \\
(5)    & \textbf{PAC}                              & 63.15                                  & 69.30                                  & 40.68                                  & 86.36                                  &  9.58                                  & 269.07                                                \\
(6)    & \textbf{MVC}                              & 63.30                                  & 68.32                                  & 40.94                                  & 85.99                                  & 10.83                                  & 269.38                                                \\
(7)    & \textbf{ITC}                              & \cellcolor[HTML]{EFEFEF}64.63          & \cellcolor[HTML]{C0C0C0}70.61          & 43.13                                  & 86.25                                  & 10.69                                  & 275.31                                                \\ \hdashline
(8)    & \textbf{ITC + MLM + MPFC}                 & 64.28                                  & 70.02                                  & 43.31                                  & \cellcolor[HTML]{C0C0C0}87.21                                  & \cellcolor[HTML]{EFEFEF}11.12                                  & 275.94                                                \\
(9)    & \textbf{ITC + MLM + MPFC + ITM}           & 64.37                                  & \cellcolor[HTML]{EFEFEF}70.44                                  & \cellcolor[HTML]{EFEFEF}43.56                                  & \cellcolor[HTML]{EFEFEF}87.17          & 11.08                                  & \cellcolor[HTML]{EFEFEF}276.62                                                \\
\,(10)\,   & \textbf{ITC + MLM + MPFC + ITM + MVC}           & \cellcolor[HTML]{C0C0C0}64.88                                  & 70.34          & \cellcolor[HTML]{C0C0C0}43.94          & 87.12          & \cellcolor[HTML]{C0C0C0}11.56          & \cellcolor[HTML]{C0C0C0}{\color[HTML]{343434} 277.84} \\
\,(11)\,   & \textbf{ITC + MLM + MPFC + ITM + MVC + PAC}           & \cellcolor[HTML]{9B9B9B}65.00 & \cellcolor[HTML]{9B9B9B}70.63 & \cellcolor[HTML]{9B9B9B}44.12 & \cellcolor[HTML]{9B9B9B}87.63 & \cellcolor[HTML]{9B9B9B}11.98 & \cellcolor[HTML]{9B9B9B}279.36               \\ \hdashline
\specialrule{0em}{3pt}{3pt}
\,(12)\,    & \makecell[l]{\textbf{ITC + MLM + MPFC + ITM + MVC + PAC} \\ (w/o sharing TE and FE)}                              & 64.16                    & 69.15                               & 42.87                           &  86.22                               & 11.31                                  &  273.71                                               \\ \hline
\end{tabular}
}
\end{table}

%% file: Figures/4-tsne_embeddings.tex
\begin{figure}[t]
\centering
\subfigure{
\includegraphics[width=0.475\linewidth]{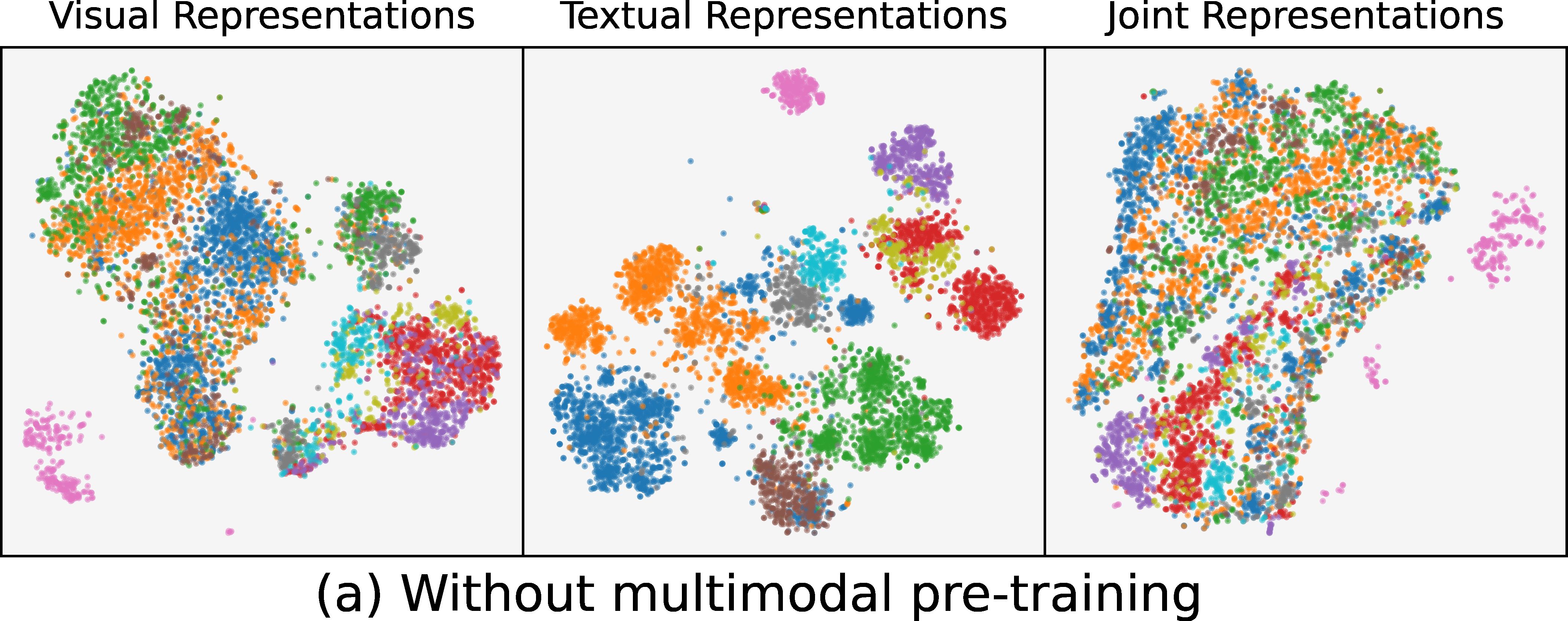}
}
\subfigure{
\includegraphics[width=0.475\linewidth]{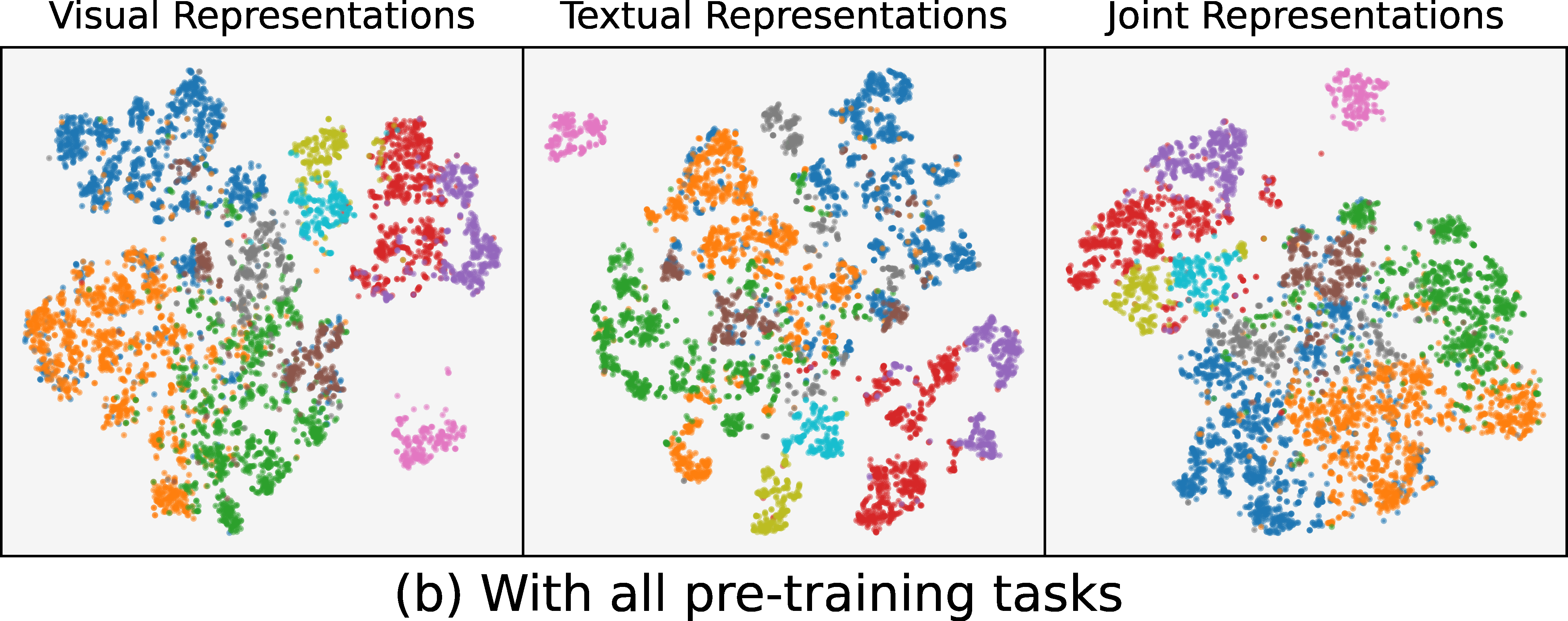}
}
\caption{T-sne of the learned visual/textual/joint representations from FashionViL}
\label{fig:tsne_embeddings}
\end{figure}

%% file: Sections/5-conclusion.tex
\section{Conclusions}
We have introduced FashionViL, a novel end-to-end large-scale pre-training framework for V+L representation learning in the fashion domain.
We proposed two effective fashion-specific pre-training tasks and introduced a novel modality-agnostic text/fusion encoder for a flexible and versatile multimodal architecture.
Our FashionViL achieves new SOTA performance with superior efficiency on 5 popular fashion-related tasks.

%% file: Sections/a-supp.tex
\appendix
This supplementary material includes three sections.
Sec. \ref{apx:implementation_details} describes our implementation details for the pre-training pipeline and each downstream task.
Sec. \ref{apx:additional_results} shows more experiments to demonstrate the effectiveness of FashionViL.
Sec. \ref{apx:additional_vis} provides the additional visualization examples.

\section{Implementation details}
\label{apx:implementation_details}

\subsection{Pre-training}
\statement{Image tokenizer.}
As discussed in the main paper, we adopt the Masked Patch Feature Classification (MPFC) as one of our pre-training tasks.
An image tokenizer is used to convert the raw pixel values into discrete labels.
While previous works like BEiT~\cite{bao2021beit} applied the off-the-shelf image tokenizer pre-trained on the large-scale generic image data~\cite{ramesh2021dalle}, we train the image tokenizer by ourselves on the four available fashion datasets~\cite{rostamzadeh2018fashiongen,yang2020facad,vasileva2018polyvoreoutfits,han2017fashion200k} as focusing more on the fashion domain.

Specifically, we implement a vector-quantized VAE (VQVAE)~\cite{van2017vqvae} with similar Encoder and Decoder architectures as VQGAN~\cite{esser2021taming}.
The model details are listed in Table \ref{tab:vqaearchitecture}.
We apply the perceptual loss \cite{johnson2016perceptual} to learn the codebook, but disregard the adversarial loss which was used in VQGAN \cite{esser2021taming} as it has been shown to be trivial for the representation learning~\cite{dong2021peco}.
We adopt the same training objective as PeCo~\cite{dong2021peco} to learn our VQVAE with the hyper-parameters listed in Table~\ref{tab:vqaear_hyper}.
Some reconstruction samples can be found in Fig. \ref{fig:reconstruction}.

\statement{Pre-training.}
FashionViL is end-to-end pre-trained on 6 tasks as mentioned in the main paper.
Previous fashion V+L works, \textit{i.e.} FashionBERT \cite{gao2020fashionbert} and KaleidoBERT \cite{zhuge2021kaleido}, perform all the pre-training tasks in one iteration, which is memory demanding.
In this work, we follow UNITER~\cite{chen2020uniter} to sample one task per iteration and train it with one objective. 

We implement FashionViL pre-training with MMF \cite{singh2020mmf} on 4 RTX 3090 GPUs.
All hyper-parameters are listed in Table \ref{tab:pretrain_hyper}.

\input{Tables/a0-image_tokenizer}
\input{Tables/a1-pretrain-hyper}
\clearpage

\subsection{Fine-tuning}
\statement{Cross-modal retrieval (ITR \& TIR).}
\input{Tables/a2-xmodal_hyper}
As ITR and TIR have the same objective as image-text contrastive learning (ITC), we directly fine-tune FashionViL with $\mathcal{L}_\mathrm{ITC}$ on the FashionGen dataset \cite{rostamzadeh2018fashiongen}, where the learnable temperature $\tau$ is initialized as $0.625$.
All hyper-parameters are listed in Table \ref{tab:xmodal_hyper}.

\statement{Text-guided image retrieval (TGIR).}
\input{Tables/a3-composition_hyper}
Previous works~\cite{liu2021cirr,shin2021rtic} found TGIR is a sensitive task (or dataset). 
Even a small change in the training setting can result in a quite different model performance.
For a fair and stable comparison, we keep the same experimental setting for all the experiments in Table 4 in the main paper. 
Specifically, we removed tricks like ensemble learning and only keep the composition module implementation. 
For methods with lightweight text encoders (C5, C6, C7), we use CLIP embeddings~\cite{radford2021clip} as the initialization of the word embeddings, which is shown to be effective in~\cite{han2021tbps}.
We apply batch-based classification (BBC) loss \cite{vo2019tirg} for TGIR.
All experiments are conducted using the hyper-parameters in Table \ref{tab:composition_hyper}.

\statement{Category / Subcategory recognition (CR / SCR).}
\input{Tables/a4-classification_hyper}
For CR and SCR, we directly follow the setting of KaleidoBERT \cite{zhuge2021kaleido} with the cross entropy (CE) as the loss function.
All the hyper-parameters are listed in Table \ref{tab:classification_hyper}.

\statement{Outfit complementary item retrieval (OCIR).}
\input{Tables/a5-ocir_hyper}
We follow CSA-Net~\cite{lin2020csanet} for the task of OCIR.
We tried hard but cannot get the proposed data splits and reproduction code in CSA-Net~\cite{lin2020csanet}.
We thus reorganize Polyvore Outfits \cite{vasileva2018polyvoreoutfits} and reproduce CSA-Net by ourselves according to the paper.
As a result, our results differ from the original paper, but we will release our splits and reproduction code for the convenience of future research.
All the experiments implemented by us follow the same hyper-parameters listed in Table \ref{tab:ocir_hyper}.
Contrastive loss is applied as the training objective.

\clearpage
\section{Additional quantitative results}
\label{apx:additional_results}
\subsection{Performance on multi-image
subcategory recognition}
Our model can be easily extended to support multi-image input by concatenating all image tokens together.
However, there is no existing downstream task taking multiple images for direct comparison with published results, thus such experiments are omitted. 
We have now simulated a new one -- multi-image subcategory recognition (M-SCR), which takes multiple images as input.
Table~\ref{tab:m-scr} shows that our pre-training (pt) can yield even larger gain (Acc \& Macro$\mathcal{F}$).
More interestingly, SCR outperforms M-SCR w/o pre-training, but the comparison is reversed after pre-training, indicating (a) the fusion of multiple images and text is not trivial, and (b) our FashionViL is effective in the fusion task.

\section{Additional qualitative results}
\label{apx:additional_vis}
We provide more visualization results in this section to better understand the performance of our FashionViL in a qualitative way.

\subsection{VQVAE reconstruction}
We show some reconstruction results generated by our VQVAE in Figure \ref{fig:reconstruction}.
The overall quality of the reconstructed images is satisfactory with those basic semantic information (\textit{e.g.}, the outline and color of the object) well preserved.

\subsection{Additional t-sne visualization}
We provide more t-sne visualizations for FashionViL's joint representations on the fine-grained categories in Figure \ref{fig:additional_tsne}.
In each column, we visualize all t-sne embeddings belonging to the same category (\textit{e.g.}, \texttt{TOPS}) and color them according to their subcategory labels (\textit{e.g.}, \texttt{BLOUSES} and \texttt{T-SHIRTS}).
With the help of our pre-training tasks, the multimodal representations are better clustered in the latent space at both category-level and subcategory-level, which further proved the effectiveness of our pre-training.

\input{Tables/a6-m-scr}
\clearpage
\input{Figures/a0-reconstruction}
\input{Figures/a1-additional_tsne}

%% file: Tables/a0-image_tokenizer.tex
\begin{table}[t]
\centering
\caption{High-level architecture of the encoder and decoder of our VQVAE}
\label{tab:vqaearchitecture}
\resizebox{0.6\textwidth}{!}{
\begin{tabular}{c}
\textbf{Encoder} \\
\toprule
$ x \in \mathbb{R}^{224 \times 224 \times 3}$   \\
Conv2D $\to \mathbb{R}^{224 \times 224 \times 128} $  \\
$6 \times $ $\{$Res Block, Res Block, Downsample Block$\}$ $\to \mathbb{R}^{7 \times 7 \times 512} $  \\
$2 \times $ $\{$Non-local Block, Res Block$\}$ $\to \mathbb{R}^{7 \times 7 \times 512} $ \\
GroupNorm, Swish, Conv2D $\to \mathbb{R}^{7 \times 7 \times 256}$  \\
\bottomrule
\\
\textbf{Decoder} \\ \toprule
$\ z_{\mathbf{q}} \in \mathbb{R}^{7 \times 7 \times 256}$ \\
Conv2D $\to \mathbb{R}^{7 \times 7 \times 512} $\\ $2 \times $ $\{$Res Block, Non-local Block$\}$ $\to \mathbb{R}^{7 \times 7 \times 512} $ \\
$6 \times $ $\{$Res Block, Res Block, Upsample Block$\}$ $\to \mathbb{R}^{7 \times 7 \times 128} $ \\
GroupNorm, Swish, Conv2D $\to \mathbb{R}^{224 \times 224 \times 3}$\\ \bottomrule
\end{tabular}
}
\end{table}

\begin{table}[t]
\centering
\caption{Hyper-parameters for training our VQVAE}
\label{tab:vqaear_hyper}
\resizebox{0.65\textwidth}{!}{
\begin{tabular}{llc}
\toprule
\textbf{Data augmentation}                    & RandomResizedCrop        & (224, 224)      \\ \hline
\multirow{3}{*}{\textbf{Model configuration\,}} & Codebook size            & 1024            \\
                                              & Latent feature dimension & 256             \\
                                              & EMA decay                & 0.99            \\ \hline
\multirow{4}{*}{\textbf{Training setting}}    & Number of iterations     & 500,000         \\
                                              & Batch size               & 32              \\
                                              & Initial LR               & 1.44e-4         \\
                                              & Optimizer                & Adam (0.5, 0.9) \\ \hline
\multirow{2}{*}{\textbf{Hardware}}            & GPU                      & 4 x RTX 3090    \\
                                              & Training duration        & 96h             \\ \bottomrule
\end{tabular}
}
\end{table}

%% file: Tables/a1-pretrain-hyper.tex
\begin{table}[t]
\centering
\caption{Hyper-parameters for pre-training FashionViL}
\label{tab:pretrain_hyper}
\resizebox{0.65\textwidth}{!}{
\begin{tabular}{llc}
\toprule
\textbf{Image encoder}                      &  & ResNet50           \\ \hline
\textbf{Text/Fusion encoder \,}                &  & BERT-base-uncased  \\ \hline
\multirow{3}{*}{\textbf{Text tokenizer}}    & Sequence length      & 75                 \\
                                            & Mask probability     & 15\%               \\
                                            & Whole word mask      & \checkmark                \\ \hline
\multirow{3}{*}{\textbf{Image tokenizer}}   & Min masked patches   & 4                  \\
                                            & Max masked patches   & 8                  \\
                                            & Aspect ratio of mask & (1/3, 3)           \\ \hline
\multirow{3}{*}{\textbf{Data augmentation}} & Resize               & (256, 256)         \\
                                            & RandomCrop           & (224, 224)         \\
                                            & RandomHorizontalFlip & \checkmark                \\ \hline
\multirow{11}{*}{\textbf{Training setting}} & Number of iterations & 120,000            \\
                                            & Batch size           & 256                \\
                                            & Initial LR of TE/FE  & 1e-5               \\
                                            & Initial LR of  IE    & 2e-4               \\
                                            & LR schedule          & Multi-step         \\
                                            & LR steps             & 45,000 and 90,000  \\
                                            & LR decrease ratio    & 0.1                \\
                                            & Warmup iterations    & 15,000             \\
                                            & Warmup factor        & 0.25               \\
                                            & Optimizer            & AdamW (0.9, 0.999) \\
                                            & Weight decay         & 1e-4               \\ \hline
\multirow{2}{*}{\textbf{Hardware}}          & GPU                  & 4 $\times$ RTX 3090       \\
                                            & Training duration    & 28.5h              \\ \bottomrule
\end{tabular}
}
\end{table}

%% file: Tables/a2-xmodal_hyper.tex
\begin{table}[t]
\centering
\caption{Hyper-parameters for fine-tuning FashionViL on cross-modal retrieval}
\label{tab:xmodal_hyper}
\resizebox{0.65\textwidth}{!}{
\begin{tabular}{llc}
\toprule
\textbf{Image encoder}                      &   & ResNet50           \\ \hline
\textbf{Text/Fusion encoder}                &   & BERT-base-uncased  \\ \hline
\textbf{Text tokenizer}                     & Sequence length      & 75                 \\ \hline
\multirow{3}{*}{\textbf{Data augmentation}} & Resize               & (256, 256)         \\
                                            & RandomCrop           & (224, 224)         \\
                                            & RandomHorizontalFlip & \checkmark                \\ \hline
\multirow{11}{*}{\textbf{Training setting}} & Number of iterations & 75,120             \\
                                            & Batch size           & 64                 \\
                                            & Initial LR of TE    & 1e-5               \\
                                            & Initial LR of  IE    & 2e-4               \\
                                            & LR schedule          & Multi-step         \\
                                            & LR steps             & 28,170 and 56,340  \\
                                            & LR decrease ratio    & 0.1                \\
                                            & Warmup iterations    & 9,390               \\
                                            & Warmup factor        & 0.25               \\
                                            & Optimizer            & AdamW (0.9, 0.999) \\
                                            & Weight decay         & 1e-4               \\ \hline
\multirow{2}{*}{\textbf{Hardware}}          & GPU                  & 1 x RTX 3090       \\
                                            & Training duration    & 9h                 \\ \bottomrule
\end{tabular}
}
\end{table}

%% file: Tables/a3-composition_hyper.tex
\begin{table}[t]
\centering
\caption{Hyper-parameters for fine-tuning FashionViL on TGIR}
\label{tab:composition_hyper}
\resizebox{0.65\textwidth}{!}{
\begin{tabular}{llc}
\toprule
\textbf{Image encoder}                      &   & ResNet50           \\ \hline
\textbf{Text/Fusion encoder}                &   & BERT-base-uncased  \\ \hline
\textbf{Text tokenizer}                     & Sequence length      & 75                 \\ \hline
\multirow{3}{*}{\textbf{Data augmentation}} & Resize               & (256, 256)         \\
                                            & RandomCrop           & (224, 224)         \\
                                            & RandomHorizontalFlip & \checkmark                \\ \hline
\multirow{11}{*}{\textbf{Training setting}} & Number of iterations & 44,960             \\
                                            & Batch size           & 32                 \\
                                            & Initial LR of FE  & 1e-5               \\
                                            & Initial LR of  IE    & 2e-4               \\
                                            & LR schedule          & Multi-step         \\
                                            & LR steps             & 16,860 and 28,100  \\
                                            & LR decrease ratio    & 0.1                \\
                                            & Warmup iterations    & 2,810               \\
                                            & Warmup factor        & 0.25               \\
                                            & Optimizer            & AdamW (0.9, 0.999) \\
                                            & Weight decay         & 1e-4               \\ \hline
\multirow{2}{*}{\textbf{Hardware}}          & GPU                  & 1 x RTX 3090       \\
                                            & Training duration    & 5.5h               \\ \bottomrule
\end{tabular}
}
\end{table}

%% file: Tables/a4-classification_hyper.tex
\begin{table}[t]
\centering
\caption{Hyper-parameters for fine-tuning FashionViL on (S)CR}
\label{tab:classification_hyper}
\resizebox{0.65\textwidth}{!}{
\begin{tabular}{llc}
\toprule
\textbf{Image encoder}                      &   & ResNet50           \\ \hline
\textbf{Text/Fusion encoder}                &   & BERT-base-uncased  \\ \hline
\textbf{Text tokenizer}                     & Sequence length      & 75                 \\ \hline
\multirow{3}{*}{\textbf{Data augmentation}} & Resize               & (256, 256)         \\
                                            & RandomCrop           & (224, 224)         \\
                                            & RandomHorizontalFlip & \checkmark                \\ \hline
\multirow{9}{*}{\textbf{Training setting}}  & Number of iterations & 37,580             \\
                                            & Batch size           & 32                 \\
                                            & Initial LR of FE  & 1e-5               \\
                                            & Initial LR of  IE    & 2e-4               \\
                                            & Optimizer            & AdamW (0.9, 0.999) \\
                                            & Weight decay         & 1e-4               \\ \hline
\multirow{2}{*}{\textbf{Hardware}}          & GPU                  & 1 x RTX 3090       \\
                                            & Training duration    & 2.5h               \\ \bottomrule
\end{tabular}
}
\end{table}

%% file: Tables/a5-ocir_hyper.tex
\begin{table}[t]
\centering
\caption{Hyper-parameters for fine-tuning FashionViL on OCIR}
\label{tab:ocir_hyper}
\resizebox{0.65\textwidth}{!}{
\begin{tabular}{llc}
\toprule
\textbf{Image encoder}                      &   & ResNet50           \\ \hline
\multirow{3}{*}{\textbf{Data augmentation}} & Resize               & (256, 256)         \\
                                            & RandomCrop           & (224, 224)         \\
                                            & RandomHorizontalFlip & \checkmark                \\ \hline
\multirow{9}{*}{\textbf{Training setting}}  & Number of iterations & 8,000              \\
                                            & Batch size           & 64                 \\
                                            & Initial LR of  IE    & 1e-4               \\
                                            & LR schedule          & Multi-step         \\
                                            & LR steps             & 1,500 and 5,000  \\
                                            & LR decrease ratio    & 0.1                \\
                                                                    & Warmup iterations    & 1,000               \\
                                            & Warmup factor        & 0.25               \\
                                            & Optimizer            & AdamW (0.9, 0.999) \\
                                            & Weight decay         & 1e-4               \\ \hline
\multirow{2}{*}{\textbf{Hardware}}          & GPU                  & 1 x RTX 3090       \\
                                            & Training duration    & 1.5h               \\ \bottomrule
\end{tabular}
}
\end{table}

%% file: Tables/a6-m-scr.tex
\begin{table}[t]
\centering
\caption{Results of multi-image
subcategory recognition on FashionGen \cite{rostamzadeh2018fashiongen}}
\label{tab:m-scr}
\setlength\tabcolsep{8pt}
\resizebox{0.7\linewidth}{!}{
\begin{tabular}{cccc|cccc}
\hline
\multicolumn{2}{c}{\textbf{SCR w/o pt}} & \multicolumn{2}{c|}{\textbf{SCR w/ pt}} & \multicolumn{2}{c}{\textbf{M-SCR w/o pt}} & \multicolumn{2}{c}{\textbf{M-SCR w/ pt}} \\ \hline
91.45                & 78.13               & 92.33               & 83.02               & 90.33               & 72.16               & 93.39               & 84.30              \\ \hline
\end{tabular}
}
\end{table}

%% file: Figures/a0-reconstruction.tex
\begin{figure}[t]
\begin{center}
\includegraphics[width=\linewidth]{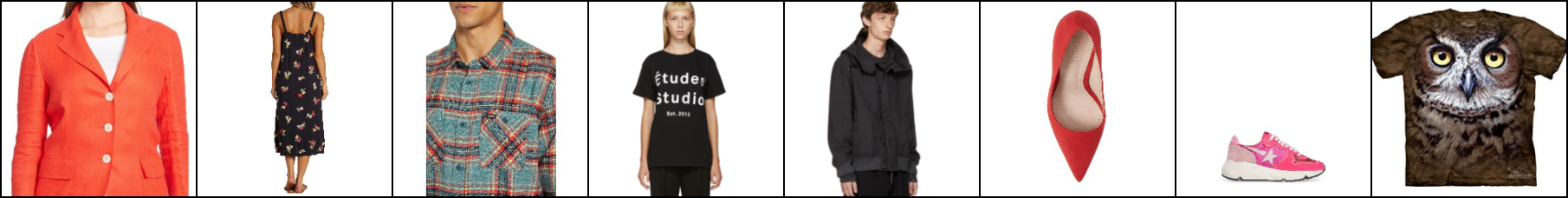}
\includegraphics[width=\linewidth]{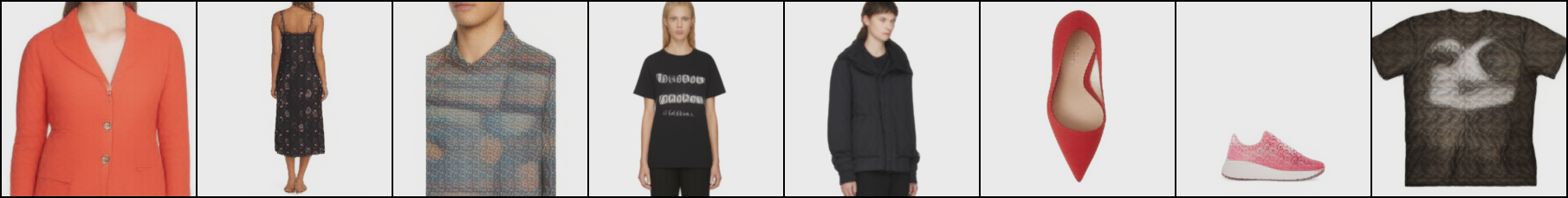}
\includegraphics[width=\linewidth]{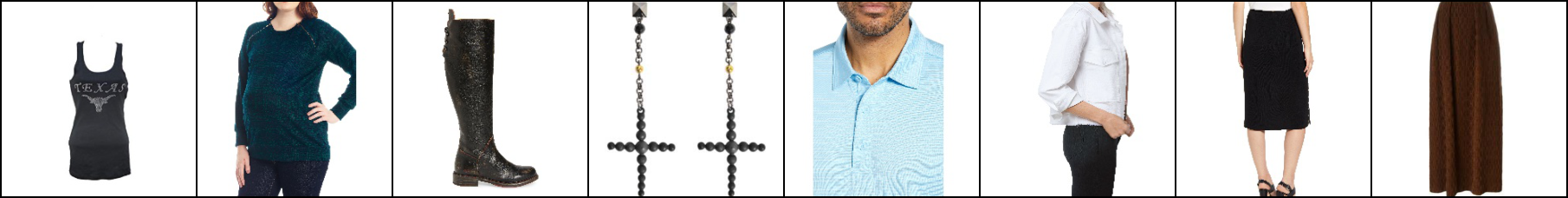}
\includegraphics[width=\linewidth]{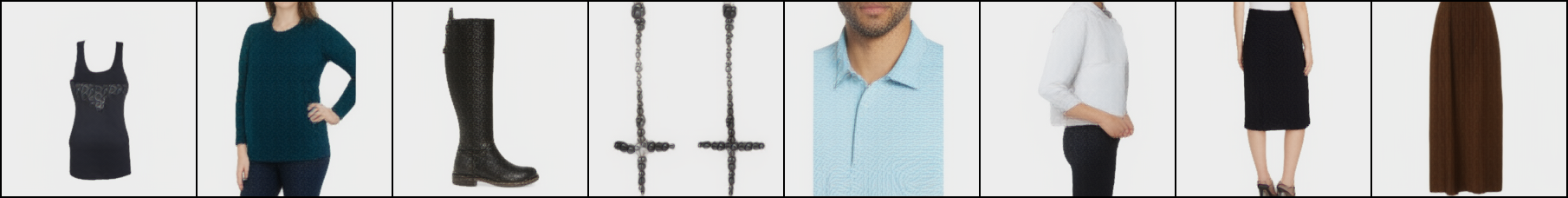}
\includegraphics[width=\linewidth]{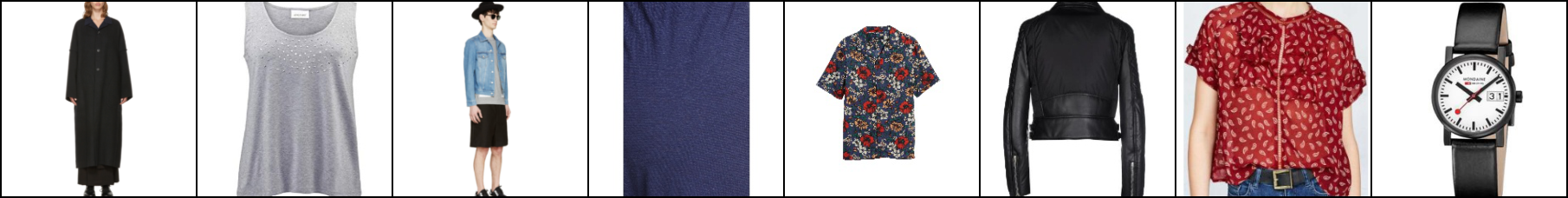}
\includegraphics[width=\linewidth]{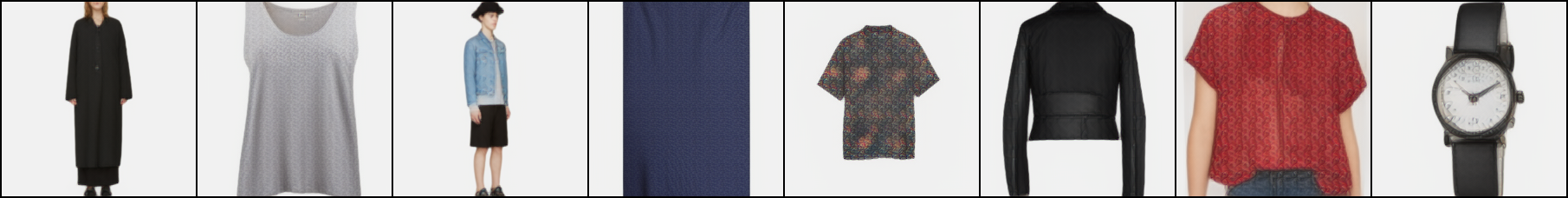}
\end{center}
\caption{Some reconstruction results generated by our VQVAE. Odd rows are the original images, and even rows are the reconstructed images from the previous row}
\label{fig:reconstruction}
\end{figure}

%% file: Figures/a1-additional_tsne.tex
\begin{figure}[t]
\centering
\subfigure{
\includegraphics[width=\linewidth]{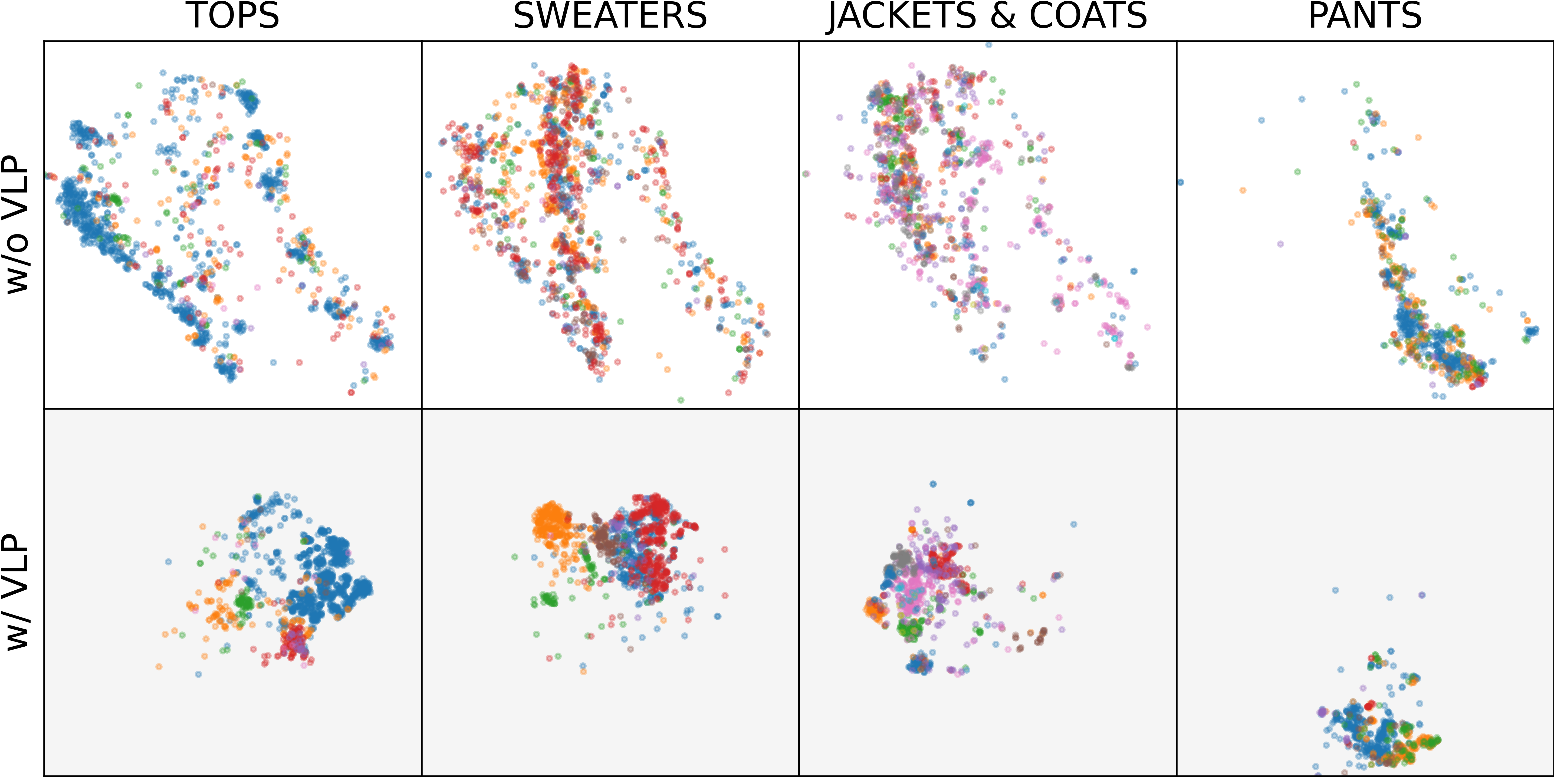}
}
\caption{T-sne of the multimodal representations from not pre-trained and pre-trained FashionViL.
Different colors represent subcategories of the categories mentioned in each column header
}
\label{fig:additional_tsne}
\end{figure}